\documentclass[letterpaper, 10 pt, conference]{ieeeconf}  % Comment this line out if you need a4paper

\IEEEoverridecommandlockouts                              % This command is only needed if 
\usepackage{amsmath,amssymb,amsfonts}
\usepackage{algorithmic}
\usepackage{graphicx}
\usepackage{comment}
\usepackage{subcaption}
\usepackage{algorithm,algorithmic}

\title{\LARGE \bf
An AI Framework for Microanastomosis Motion Assessment
}

\author{Yan Meng$^{1}$, Eduardo J. Torres-Rodríguez$^{1,2}$, Marcelle Altshuler$^{3}$, Nishanth Gowda$^{4}$, Arhum Naeem$^{4}$, \\ Recai Yilmaz$^{1}$, Omar Arnaout$^{3}$ and Daniel A. Donoho $^{1}$% <-this % stops a space
	\thanks{$^{1}$ \raggedright Yan Meng, Recai Yilmaz, and Daniel Donoho are with Department of Neurosurgery, Children's National Hospital, Washington, DC 20010, USA 
	        {\tt\small \{ymeng,myilmaz,ddonoho\}@childrensnational.org}}%
	\thanks{$^{2}$ \raggedright Eduardo J. Torres-Rodríguez was with Children's National Hospital. He is 
		now with Lombardi Comprehensive Cancer Center, Georgetown University Medical Center, Washington, DC 20007, USA
	        {\tt\small et739@georgetown.edu}}%
	\thanks{$^{3}$ \raggedright Marcelle Altshuler and Omar Arnaout are with Department of Neuro- surgery, Brigham and Women’s Hospital, Harvard Medical School, Boston, MA 02115, USA \newline
	 	{\tt\small \{maltshuler,oarnaout\}@bwh.harvard.edu }}%
	\thanks{$^{4}$ \raggedright Nishanth Gowda and Arhum Naeem are with School of Medicine and Health Sciences, George Washington University, Washington, DC 20052, USA
		{\tt\small \{nishbg16,anaeem\}@gwu.edu}}%
}

\begin{document}

\maketitle
\thispagestyle{empty}
\pagestyle{empty}

%%%%%%%%%%%%%%%%%%%%%%%%%%%%%%%%%%%%%%%%%%%%%%%%%%%%%%%%%%%%%%%%%%%%
\begin{abstract}

Proficiency in microanastomosis is a fundamental competency across multiple microsurgical disciplines, especially in neurosurgical training for procedures such as cerebrovascular bypass, where restoring brain perfusion and function is vital. These procedures demand exceptional precision and refined technical skills, making effective, standardized assessment methods essential for guiding surgical education and ensuring practitioner competence. Traditionally, the evaluation of microsurgical techniques has relied heavily on the subjective judgment of expert raters. While expert assessments offer valuable insights, they are inherently constrained by limitations such as inter-rater variability, lack of standardized evaluation criteria, susceptibility to cognitive bias, and the time-intensive nature of manual review. These shortcomings underscore the urgent need for an objective, reliable, and automated system capable of assessing microsurgical performance with consistency and scalability. To bridge this gap, we propose a novel AI framework for the automated assessment of microanastomosis instrument handling skills. The system integrates four core components: (1) an instrument detection module based on the You Only Look Once (YOLO) architecture; (2) an instrument tracking module developed from Deep Simple Online and Realtime Tracking (DeepSORT); (3) an instrument tip localization module employing shape descriptors; and (4) a supervised classification module trained on expert-labeled data to evaluate instrument handling proficiency. Experimental results demonstrate the effectiveness of the framework, achieving an instrument detection precision of 97\%, with a mean Average Precision (mAP) of 96\%, measured by Intersection over Union (IoU) thresholds ranging from 50\% to 95\%  (mAP50–95). Additionally, the supervised skill classification module attains an overall accuracy of 87\% in reproducing expert assessments across three categories: Poor, Moderate, and Good. These findings underscore the potential of advanced computer vision techniques to provide objective, efficient, and high-fidelity assessments of surgical skill, thereby supporting more standardized training and evaluation in neurosurgical education.
\newline

\indent \textit{Clinical relevance}—The proposed AI-aided surgical skill assessment framework facilitates an enhanced microanastomosis training experience for neurosurgeons.

\end{abstract}

%%%%%%%%%%%%%%%%%%%%%%%%%%%%%%%%%%%%%%%%%%%%%%%%%%%%%%%%%%%%%%%%%%%%%%%%%%%%%%%%
\section{INTRODUCTION}

Microsurgical competence, particularly in procedures such as microanastomosis, is fundamental to neurosurgical practice. The technical complexity of these procedures necessitates not only precise motor control but also the ability to manipulate delicate instruments within confined operative fields. Consequently, the development of microsurgical proficiency is a critical component of neurosurgical education, requiring rigorous and continuous assessment to ensure patient safety and optimal clinical outcomes \cite{reznick2006teaching,mcgoldrick2015motion}.

Traditionally, the evaluation of microsurgical skill has relied on the subjective judgment of experienced surgical educators or expert raters. While this approach provides nuanced, experience-based feedback, it is inherently limited by issues such as inter-rater variability, cognitive bias, and inconsistency in evaluation standards \cite{martin1997objective,aoun2015pilot}. Furthermore, the assessment process is often time-consuming, placing a significant burden on faculty and hindering the scalability of training programs. These limitations have spurred growing interest in the development of objective, standardized, and automated assessment tools that can deliver consistent and reproducible evaluations \cite{zia2016automated, meng2023automatic}.

Recent advances in computer vision and artificial intelligence (AI) have opened new possibilities for the quantitative analysis of surgical performance. Systems employing deep learning for instrument detection, tracking, and motion analysis have demonstrated promising results in various surgical domains, including laparoscopy, endoscopy, and robotic surgery \cite{twinanda2016endonet, funke2019video,mascagni2022artificial}. These technologies have the potential to transform surgical education by providing real-time, objective feedback and enabling data-driven assessment at scale.

In this study, we present an AI framework designed to assess instrument handling skills during microanastomosis. The key contributions of this study are:

\begin{figure*}[t]
	\centering
	\includegraphics[width=0.79\textwidth]{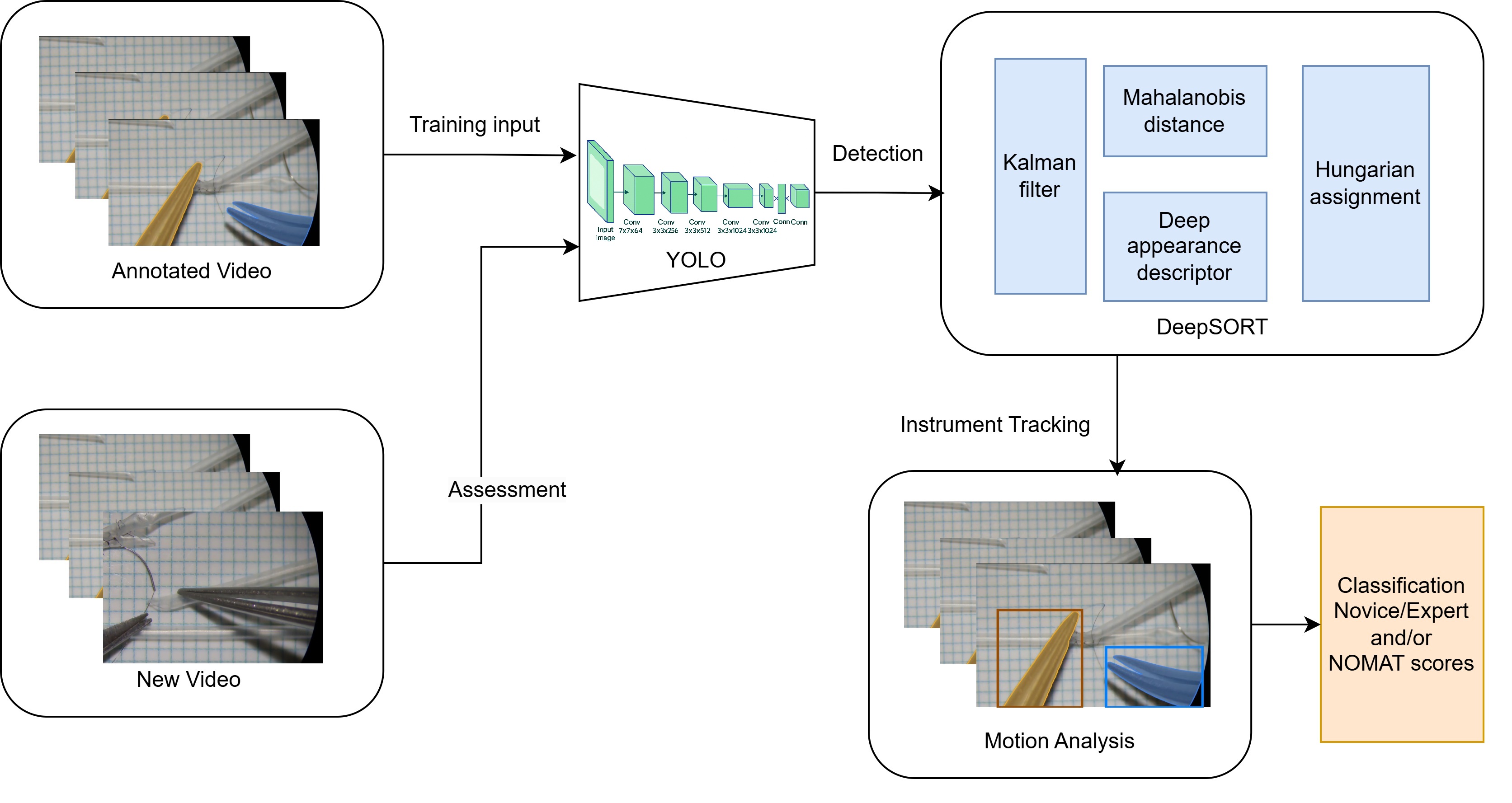}
	\caption{An overview of the AI microanastomosis motion assessment system architecture.}
	\label{fig:framework-overview}
\end{figure*}

\begin{itemize}
	\item Develop an end-to-end AI framework for assessing microanastomosis instrument handling skills, tailored to the demands of high-precision microsurgery.
	\item Design an enhanced instrument detection and tracking algorithm based on modifications to the You Only Look Once (YOLO) and Deep Simple Online and Realtime Tracking (DeepSORT) models, improving robustness and temporal accuracy.
	\item Integrate tip localization using shape descriptors to capture accurate kinematics data.
	\item Train a supervised classification model using expert-labeled data and motion feature extraction to objectively grade instrument handling proficiency, providing a scalable alternative to human evaluation.
\end{itemize}

\begin{figure*}[h]
	\centering
	\resizebox{0.9\textwidth}{!}{
		\begin{subfigure}[b]{0.46\textwidth}
			\includegraphics[width=\textwidth]{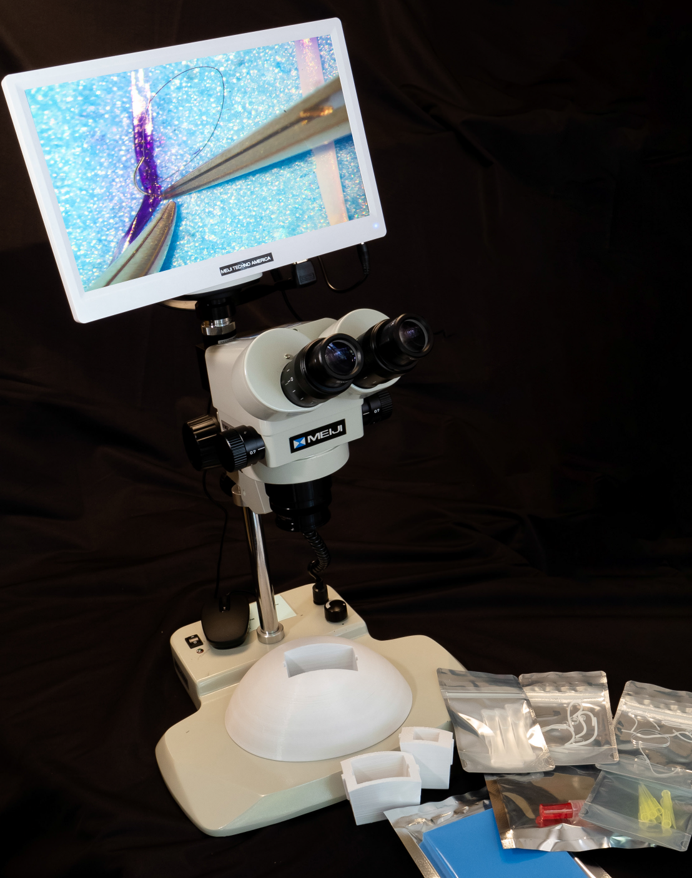}
			\caption{EMZ-250TR trinocular zoom stereo system}
			\label{fig:microscope}	
		\end{subfigure}  % -> Figure 1a
		\hfill
		\begin{minipage}[b]{0.52\textwidth}
			\centering
			\begin{subfigure}[b]{\textwidth}
				\includegraphics[width=\textwidth]{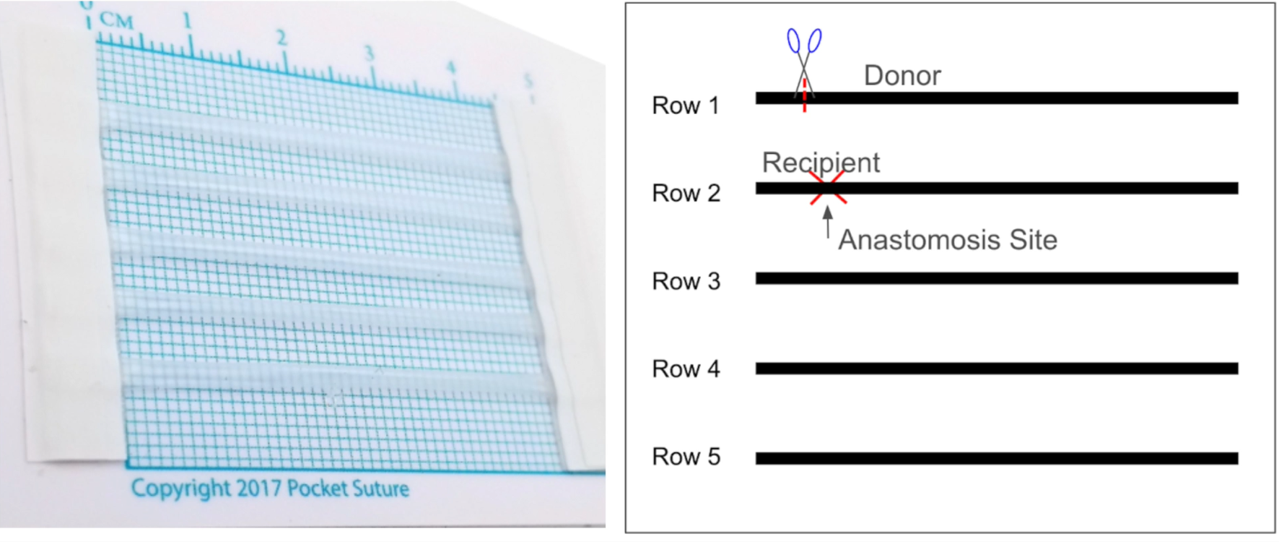}
				\caption{1.0mm $\times$ 0.8mm pocket microvascular anastomosis card}
				\label{fig:tubes}	
			\end{subfigure} 
			\vspace{0.5em}
			\begin{subfigure}[b]{\textwidth}
				\includegraphics[width=\textwidth]{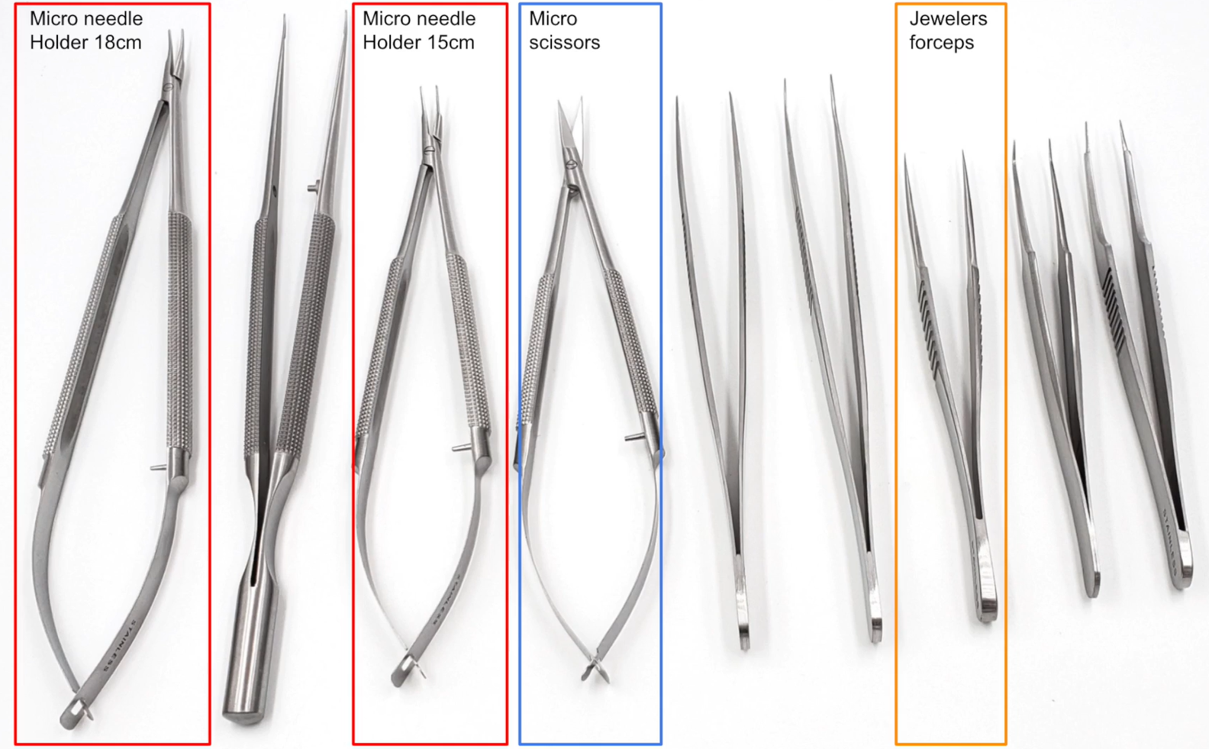}
				\caption{Microanastomosis instruments}
				\label{fig:tools}
			\end{subfigure}  % -> Figure 1c
		\end{minipage}	
	}
	\caption{Data collection and video recording environment setup.}
	\label{fig:record}
\end{figure*}

\section{Methodology}
We present a novel automated skill assessment framework for microanastomosis suturing training. The system consist of four modules: a deep learning instrument tracking module based on You Only Look Once (YOLO)\cite{redmon2016you}, an instrument tracking module developed from Deep Simple Online and Realtime Tracking (DeepSORT) \cite{wojke2017simple}, an instrument tip localization module, and a medical scoring module based on Northwestern Objective Microanastomosis Assessment Tool (NOMAT) \cite{aoun2015pilot}. An overview of the proposed architecture is provided in Fig. \ref{fig:framework-overview}.

\subsection{Instrument Detection}
We detect surgical instruments using a customized YOLOv11 model. In microanastomosis settings, each surgical instrument consists of two branches that open and close, resulting in significant appearance variations across frames. To enhance detection accuracy, we leverage the prior knowledge that each frame contains at most two surgical instruments, and the two are always of different types. Consequently, if the model detects multiple objects of the same class, a selection strategy in equation \ref{eq:iou} is applied: if the Intersection over Union (IoU) between bounding boxes is high, the boxes are merged; otherwise, only the branch with the highest confidence score is retained.

\begin{equation}	\label{eq:iou}
	\begin{aligned}
		B_{\text{final}} =
		\begin{cases}
			B_1 \cup B_2 & \text{if } \text{IoU}(B_1, B_2) > \tau_{\text{merge}} \\
			\arg\max\limits_{B \in \{B_1, B_2\}} C(B) & \text{if } \text{IoU}(B_1, B_2) \le \tau_{\text{merge}}
		\end{cases}
	\end{aligned}
\end{equation}

\noindent where $\tau_{\text{merge}}$ is a predefined IoU threshold set empirically, 0.7 gives the best results, and $ C(B)$ is the confidence score of bounding box $B$.

\subsection{Instrument Tracking}
YOLO is a frame-based object detection model, which can result in inconsistent object detection across consecutive frames, including significant variations in bounding box shapes and intermittent detection failures. For accurate surgical skill analysis based on continuous kinematic data, it is essential to maintain consistent and persistent detection of surgical instruments throughout the video sequence. To address this, we developed an object tracking algorithm based on DeepSORT, using a ResNet-based deep appearance descriptor for feature embedding \cite{he2016deep}. However, standard DeepSORT has two limitations: it often produces loose bounding boxes, and it frequently assigns new object identity when an object is temporarily misidentified as a new instance or lost track.

To mitigate these issues, we propose a hybrid detection-tracking strategy. When both YOLO-detected and DeepSORT-tracked bounding boxes are available, the YOLO bounding boxes are prioritized due to their higher spatial accuracy. In such cases, the historical bounding box data associated with the corresponding object ID is overwritten with the new YOLO-based detections. Both class IDs and object IDs are stored to enhance identification consistency. If the tracker assigns a new object ID but the class ID remains unchanged, the original object ID is preserved. This approach eliminates the dependency on DeepSORT’s memory-based object management, reducing memory usage and improving computational efficiency.

\subsection{Instrument Tip Localization}
To accurately identify the instrument tip within each bounding box, a shape descriptor is computed at multiple candidate points. These descriptors are compared to a reference descriptor representing the known appearance of the instrument tip. The matching is performed using cosine similarity in the descriptor space, and the point with the highest similarity is selected as the estimated tip location, , as defined in equation \ref{eq:tip}. This approach ensures consistent tip localization by leveraging both geometric and appearance-based features. Finally, the localized instrument tips are transformed back to the scene from the bounding box local coordinate system.

\begin{equation}	\label{eq:tip}
	\begin{aligned}
		\hat{p} = \arg\max_{i \in \{1, \dots, N\}} \cos(\theta_i) = \arg\max_{i} \frac{ \mathbf{d}_{\text{ref}} \cdot \mathbf{d}_i }{ \|\mathbf{d}_{\text{ref}}\| \|\mathbf{d}_i\| }
	\end{aligned}
\end{equation}

\noindent where $\hat{p}$ is the instrument tip; $\mathbf{d}_{\text{ref}} \in \mathbb{R}^n$ is the reference object descriptor vector; $\mathbf{d}_i \in \{\mathbb{R}^n \}_{i=1}^{N}$ is a set of feature vectors from $N$ candidate points; $\cos(\theta_p)$ is the cosine similarity between the reference and the $i_{th}$ candidate descriptor.

The complete workflow of the proposed method is outlined in Algorithm \ref*{alg:tip_tracking}, detailing each stage of the processing pipeline.

\begin{algorithm}[t]
	\caption{Instrument Tip Tracking}
	\label{alg:tip_tracking}
	\begin{algorithmic}
		\STATE
		\STATE {\textsc{PROCESS\_FRAME}}$(I_t)$
		\STATE \hspace{0.5cm}$ \mathcal{B}_t \gets \textsc{YOLO}(I_t) $ \hfill // Detect bounding boxes 
		\STATE \hspace{0.5cm} $\mathcal{B}_t = \{ \mathbf{b}_i^t \}_{i=1}^{N_t}$
		\STATE \hspace{0.5cm}$ \mathcal{T}_t \gets \textsc{DeepSort}(\mathcal{B}_t, \mathcal{T}_{t-1}) $ \hfill // Assign consistent IDs
		\STATE \hspace{0.5cm}\textbf{for each } $(\mathbf{b}_i^t, \text{id}_i^t) \in \mathcal{T}_t$ \textbf{ do}
		\STATE \hspace{1.0cm}$ \mathbf{s}_i^t \gets \textsc{ExtractShape}(\mathbf{b}_i^t) $ \hfill 
		\STATE \hspace{1.0cm}$ \mathbf{p}_i^t \gets \textsc{MatchTip}(\mathbf{s}_i^t, \Theta_{\text{prior}}) $ \hfill // Locate tip
		\STATE \hspace{1.0cm}$ \mathbf{q}_i^t \gets \textsc{TransformToScene}(\mathbf{p}_i^t, \mathbf{b}_i^t) $ \hfill 
		\STATE \hspace{0.5cm}\textbf{end for}
		\STATE \hspace{0.5cm}\textbf{return} $\{ (\text{id}_i^t, \mathbf{q}_i^t) \}_{i=1}^{N_t}$
		\STATE
		\STATE {\textsc{INITIALIZE\_SHAPE\_PRIOR}}$(\mathcal{S})$
		\STATE \hspace{0.5cm}$ \Theta_{\text{prior}} \gets \textsc{LearnShapePrior}(\mathcal{S}) $ \hfill // Fit shape model from labeled samples
		\STATE \hspace{0.5cm}\textbf{return} $\Theta_{\text{prior}}$
	\end{algorithmic}
\end{algorithm}

\section{Experiments}
We conducted an experimental study to validate the effectiveness of the proposed AI framework. The data collection and study were approved by the Institutional Review Board of Brigham and Women's Hospital.

\subsection{Data Collection}
Microanastomosis procedures were performed using simulated educational kits comprising a EMZ-250TR trinocular zoom stereo system from Meiji Techno, a HD camera with monitor, 1.0mm $\times$ 0.8mm pocket microvascular anastomosis cards by Pocket Suture, and a standardized set of instruments, including one straight needle driver, one curved needle driver, one pair each of straight and curved scissors as in Fig. \ref{fig:record}.

\begin{figure}[t]
	\centerline{\includegraphics[width=0.85\columnwidth]{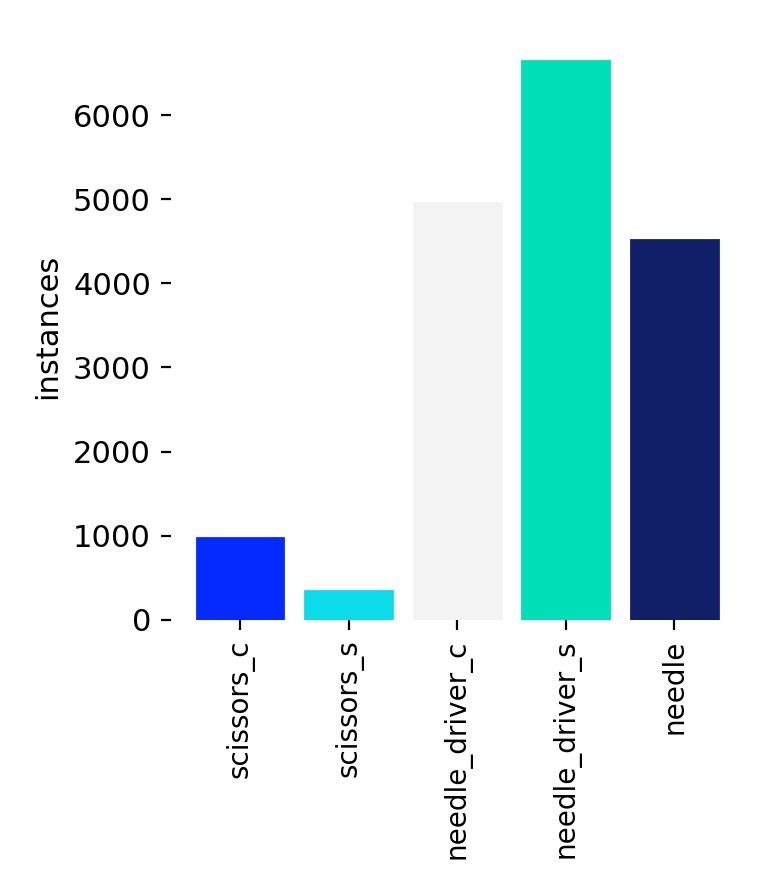}}
	\caption{Class distribution in the object detection dataset.}
	\label{fig:labels}
\end{figure}

A total of nine participants (8 male, 1 female, mean age 30.5) took part in the study, with experience ranging from novice to expert. Each participant performed between five and ten microanastomosis procedures, yielding 63 video recordings in total. Of these, 58 recordings were complete with a mean procedure length of 26 minutes and 5 were incomplete. All participants were right-handed, except for one who was ambidextrous. Each complete video captured an entire procedure, consisting of three vessel cutting actions followed by eight stitches. Four videos downsampled to 47,344 frames were used for training the object detection model, with the dataset split into 80\%, 10\%, and 10\% subsets for training, validation, and testing. Fig. \ref{fig:labels} shows the histogram of the object class distribution in the dataset. The data were annotated on Encord platform semi-automatically using Segment Anthing Model (SAM) \cite{kirillov2023segment}.

\begin{figure*}[t]
	\centering
	\begin{subfigure}{0.6\textwidth}
		\includegraphics[width=\textwidth]{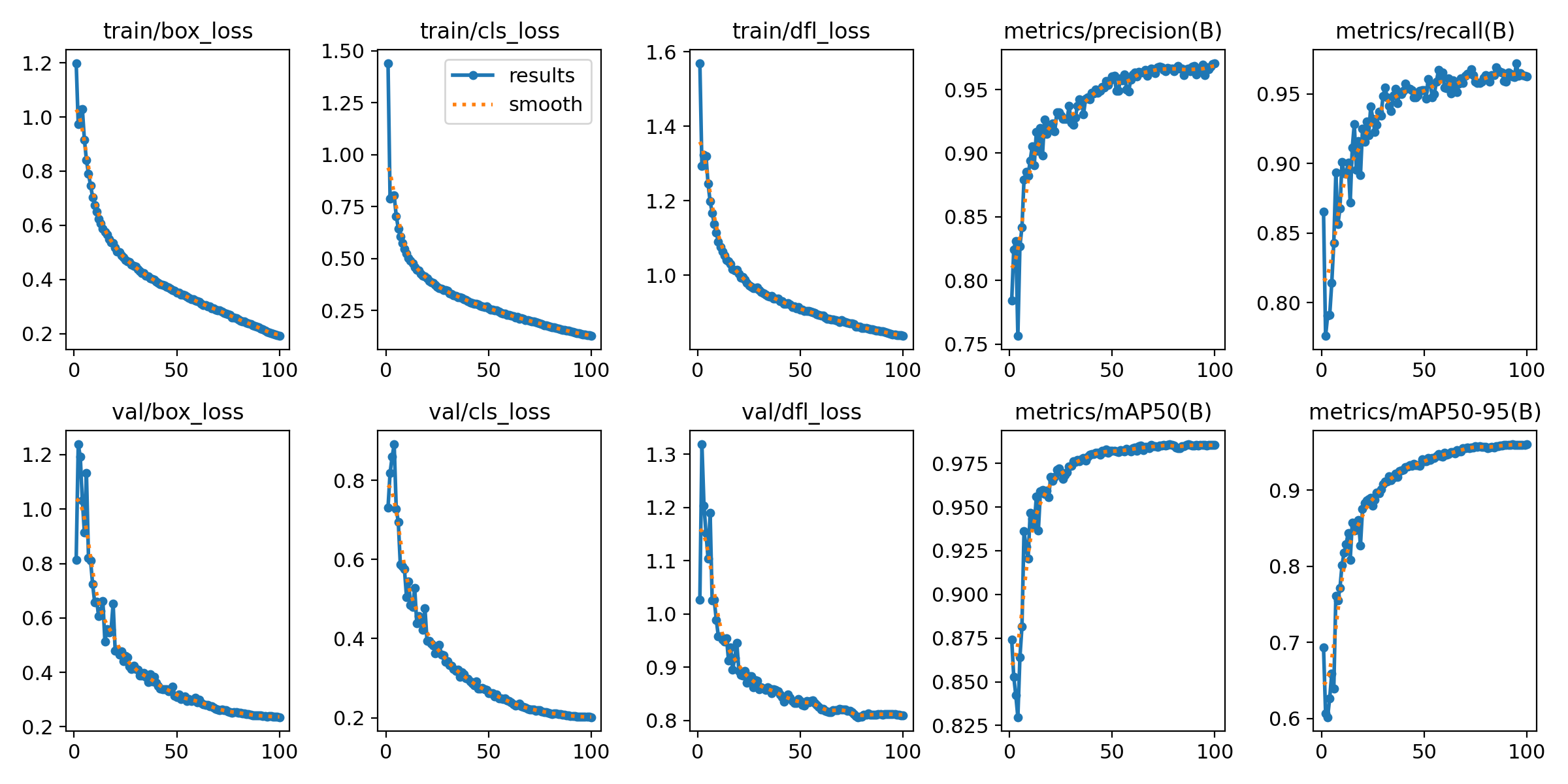}
		\caption{Loss and performance metrics convergence over epochs.}
		\label{fig:loss}
	\end{subfigure}
	\hfill
	\begin{subfigure}{0.36\textwidth}
		\includegraphics[width=\textwidth]{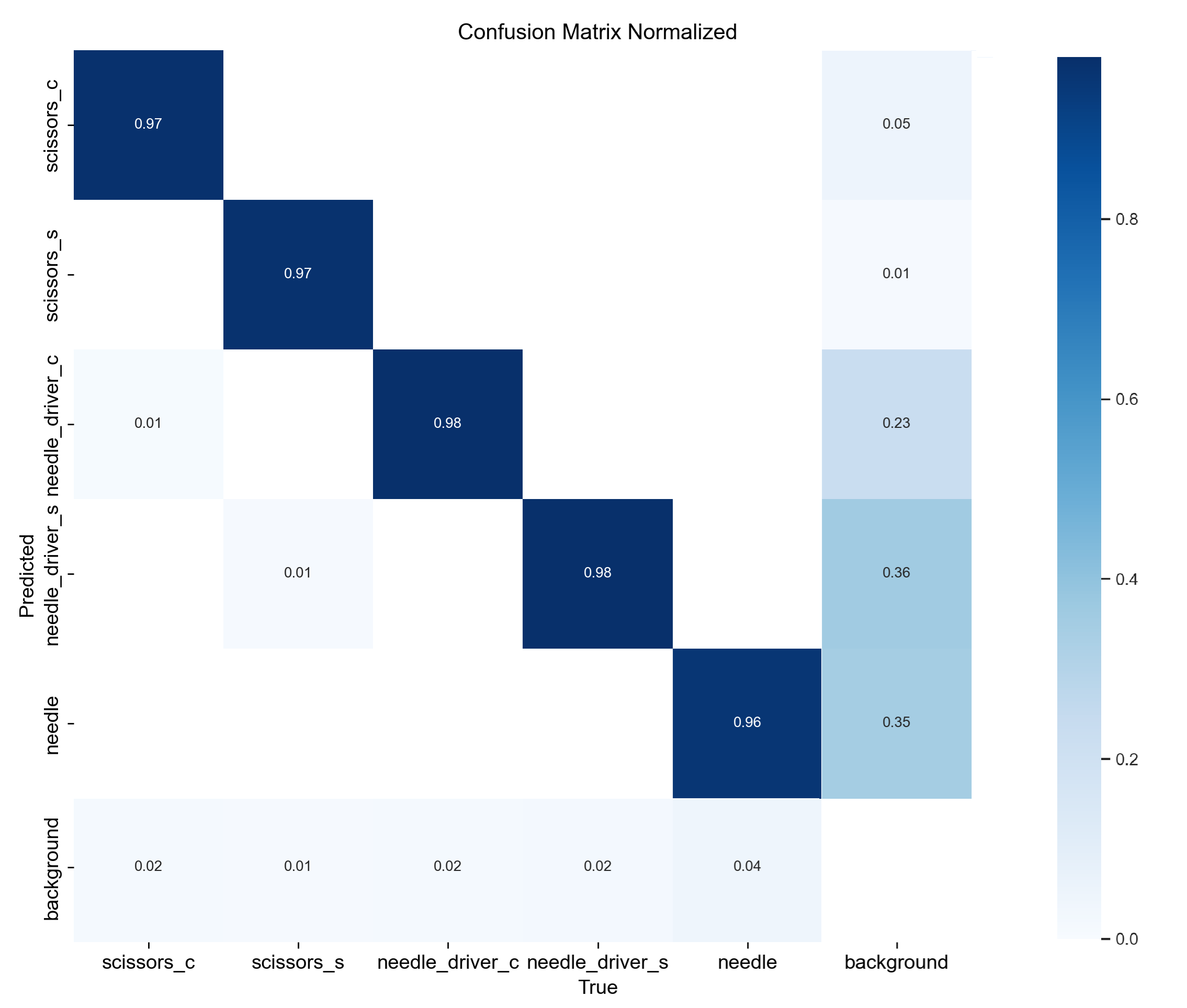}
		\caption{Normalized class confusion matrix.}
		\label{fig:confusion}
	\end{subfigure}
	\caption{YOLOv11 training results.}
\end{figure*}
\begin{figure*}[t]
	\centering
	\begin{subfigure}{0.49\textwidth}
		\includegraphics[width=\textwidth]{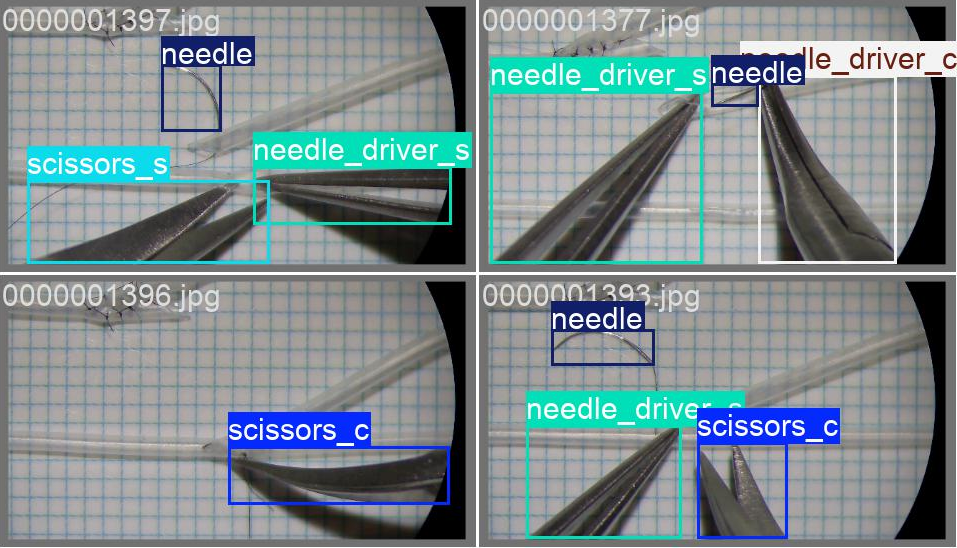}
		\caption{Ground truth labels}
		\label{fig:gt}
	\end{subfigure}
	\hfill
	\begin{subfigure}{0.49\textwidth}
		\includegraphics[width=\textwidth]{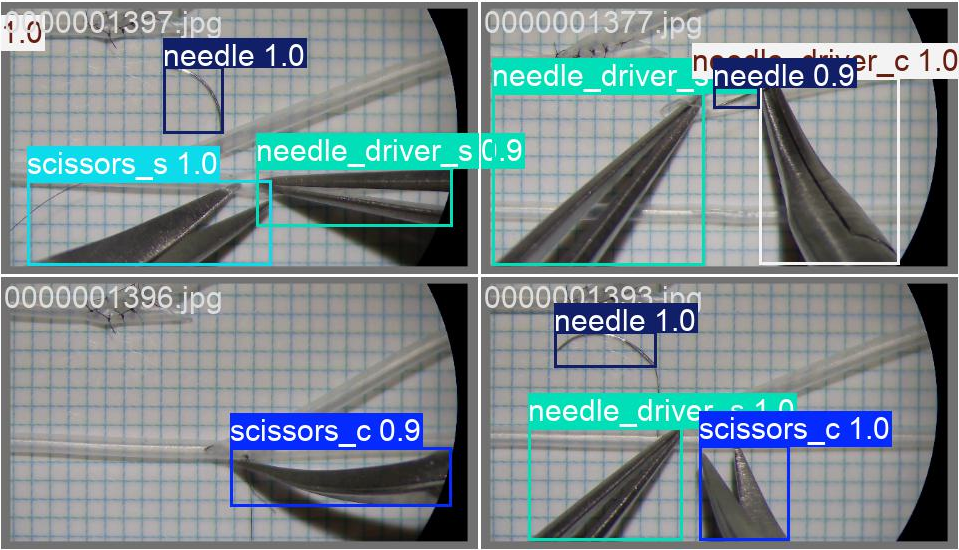}
		\caption{Predicted labels}
		\label{fig:pred}
	\end{subfigure}
	\caption{YOLOv11 testing images \label{fig:pred_image}}
\end{figure*}

\begin{figure*}[t]
	\centering
	\begin{subfigure}{0.32\textwidth}
		\includegraphics[width=\textwidth]{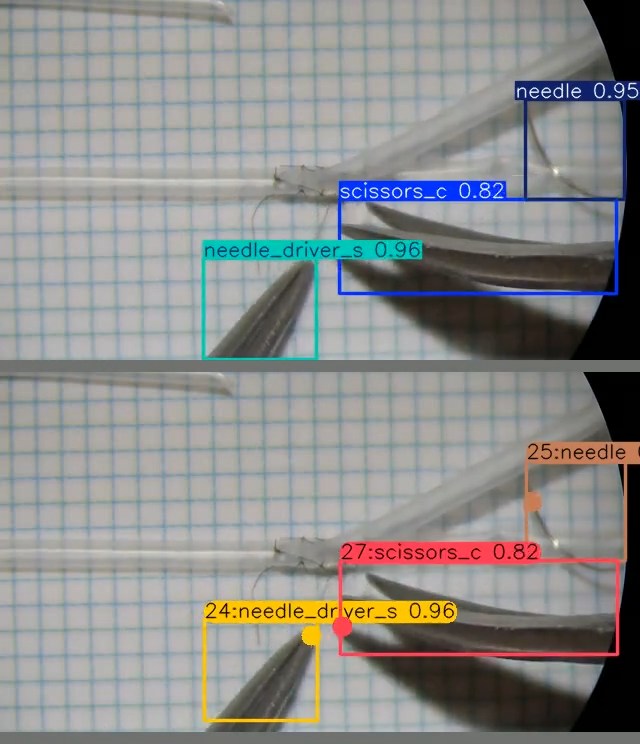}
		\caption{Participant 0824 frame 4191.}
		\label{fig:frame1}
	\end{subfigure}
	\hfill
	\begin{subfigure}{0.32\textwidth}
		\includegraphics[width=\textwidth]{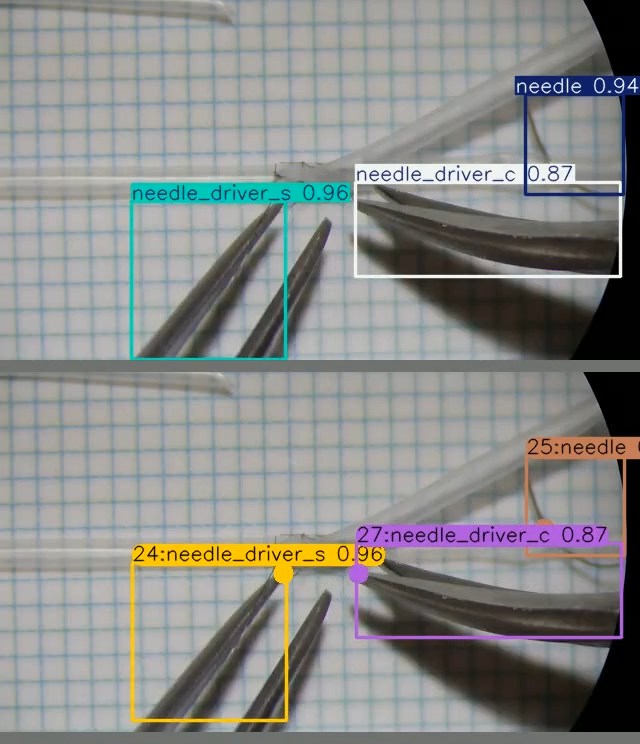}
		\caption{Participant 0824 frame 4192.}
		\label{fig:frame2}
	\end{subfigure}
	\hfill
	\begin{subfigure}{0.32\textwidth}
		\includegraphics[width=\textwidth]{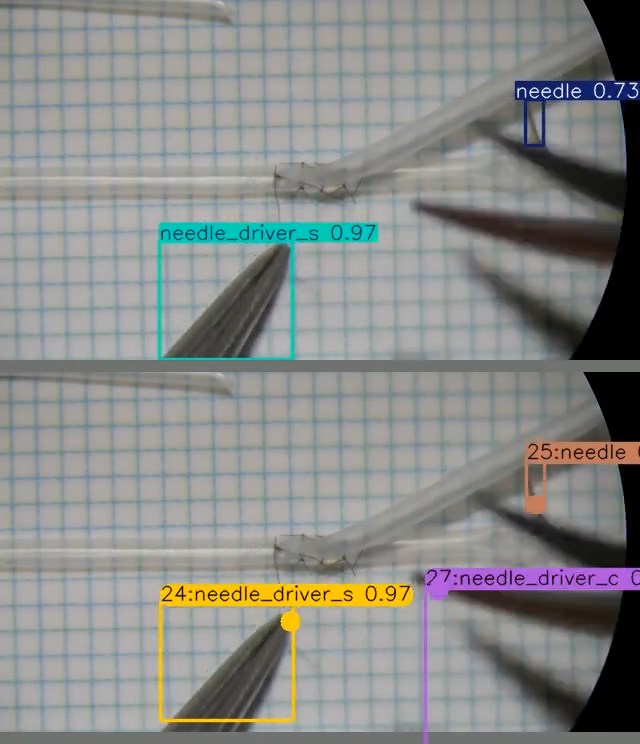}
		\caption{Participant 0824 frame 4198.}
		\label{fig:frame3}
	\end{subfigure}
	\caption{Comparison of the proposed framework with YOLOv11 results, the top row is the YOLOv11 object detection results, the bottom row is the proposed object tracking and tip localization results.\label{fig:yolo_sort}}
\end{figure*}

\begin{table*}[h]
	\caption{Instrument detection and tracking test results }
	\label{tab:val}
	\centering
	\resizebox{0.98\textwidth}{!}{
		\begin{tabular}{|c|c|c|c|c|c|c|c|c|}
			\hline
			Class    &No. of Images     & Total Instances      & Precision      & Recall  & mAP50  & mAP50–95 & Recovery	& Correction \\
			\hline
			all      & 4174	    		& 9417    	&  0.969   & 0.966  	& 0.989    & 0.958 	& 0.987	& 0.906	\\
			\hline
			scissors\_c     & 398      & 398    	& 0.981  	& 0.96		&0.992		& 0.976 	& 0.992	& 0.864\\
			\hline
			scissors\_s     & 70     & 70   		& 0.921  	& 1		&0.995		& 0.979 	 &0.995	& 0.872	\\
			\hline
			needle\_driver\_c     & 2861  & 2861	& 0.989  	& 0.975		&0.993		& 0.973 	 &0.989	& 0.902	\\
			\hline
			needle\_driver\_s     & 3669  & 3779	& 0.986  	& 0.975		&0.992		& 0.965 	 &0.994	& 0.895	\\
			\hline
			needle     & 2309  & 2309	& 0.971  	& 0.918		&0.972		& 0.872 	 & 0.963	& 0.997	\\
			\hline
		\end{tabular}
	}
\end{table*}

Two experienced neurosurgeons independently evaluated instrument handling proficiency per video using the NOMAT rubric and reached consensus on each assessment. Ratings were assigned using a five-point Likert scale, with higher scores indicating greater proficiency.

\subsection{Kinematics Classification}
We tracked the positions of instrument tips throughout each video and computed velocity, acceleration, and jerk to capture motion dynamics. To derive discriminative representations from these time series, we applied the feature extraction based on scalable hypothesis tests, which transforms variable-length temporal signals into feature vectors \cite{christ2018time}. This is achieved by computing a diverse set of statistical, frequency-domain, and model-based descriptors. The extracted features were subsequently used to train a Gradient Boosting classifier to predict procedural skill ratings \cite{konstantinov2021interpretable}. A total of 58 completed surgical procedures were used for model development, with the data split into 80\% training and 20\% testing sets. The training set was further partitioned into five folds for cross-validation, wherein the model was iteratively trained on four folds and validated on the fifth. Due to the limited size of the video dataset and the imbalanced distribution of rater scores, the scores were regrouped into three categorical classes: poor, moderate, and good, using thresholds of 2.5 and 3.5. This categorization resulted in 28, 16, and 14 samples in each class, respectively.

\section{Results}
\subsection{Instrument Detection}
The performance of the instrument detection module was evaluated using standard object detection metrics: precision, recall, and mean Average Precision (mAP). Specifically, mAP was computed at an Intersection over Union (IoU) threshold of 50\% (mAP50), and across a range from 50\% to 95\% in 5\% increments (mAP50–95). As shown in Table \ref{tab:val}, the overall detection precision across all instrument classes reached 96.9\%. The mAP50 and mAP50–95 were 98.7\% and 90.6\%, respectively. In addition, the model is lightweight and capable of real-time inference, achieving an average processing speed of 29.7 frames per second (fps).

The model was trained for 100 epochs. Fig. \ref{fig:loss} illustrates the training dynamics, showing that the model converged after approximately 50 epochs. The class-specific classification performance is visualized in the normalized confusion matrix in Fig. \ref{fig:confusion}, which shows minimal confusion between classes. Misclassifications were rare and generally occurred when instruments exited the view but cast shadows in the scene, temporarily misleading the detector. %These occurrences were infrequent in the dataset and did not significantly affect overall performance.

\begin{comment}
	The class-level analysis further validates the robustness of the framework. The curved needle driver and straight needle driver classes, which comprise the majority of the test dataset (2,861 and 3,779 instances, respectively), achieved precision values of 98.9\% and 98.6\%, with mAP50–95 values of 90.2\% and 89.5\%, respectively. The scissors and needle classes also showed strong performance, although the straight scissors class had slightly lower precision (92.1\%) due to its relatively small sample size (70 instances), which may have limited the model’s generalization.
\end{comment}

The class-level analysis further validates the robustness of the framework. The majority class, curved needle driver and straight needle driver, achieved precision values of 98.9\% and 98.6\%, with mAP50–95 values of 90.2\% and 89.5\%, respectively. The scissors and needle classes also showed strong performance, although the straight scissors class had slightly lower precision (92.1\%) due to its relatively small sample size (70 instances).

\subsection{Instrument Tracking}
Instrument tracking performance was assessed by two supplementary metrics: (1) the recovery rate, defined as the percentage of missed detections recovered by the algorithm, and (2) the correction rate, indicating how often the algorithm corrected misclassified instrument labels from the YOLO-based detector. The test results in Table \ref{tab:val} show they averaged at 98.7\% and 90.6\%, respectively, demonstrating the effectiveness of the improved tracking algorithm in mitigating YOLO's occasional misclassifications or missed detections.

Fig. \ref{fig:pred_image} presents example predictions compared with ground truth annotations, with detection confidences consistently above 90\%. Furthermore, the effect of the tracking algorithm is highlighted in Fig. \ref{fig:yolo_sort}. For instance, at frame 4191 (Fig. \ref{fig:frame1}), YOLO correctly identified a curved scissors, but in the subsequent frame 4192 (Fig. \ref{fig:frame2}), it misclassified the scissors as a curved needle driver. The object tracking module successfully corrected this by preserving the object ID across frames. Additionally, at frame 4198, the tracker recovered the scissors instance that YOLO failed to detect,and maintained a consistent object ID assignment, effectively correcting the misclassification propagated from earlier frames.

The tip localization algorithm demonstrated consistently high spatial accuracy in identifying and tracking the positions of instrument tips across sequential video frames, as shown in Fig. \ref{fig:yolo_sort}. This level of precision is essential for ensuring the reliability of downstream computational tasks, particularly those involving quantitative motion analysis, temporal kinematic profiling, and automated skill assessment.

A limitation of this method is that position estimation is based solely on 2D information, making it dependent on the camera viewpoint and unable to fully capture 3D motion.

\subsection{Skills Classification}
The instrument handling skill classifier achieves an overall accuracy of 87\% in reproducing expert assessments across three categories: Poor, Moderate, and Good, indicating robust general performance. However, a closer examination of class-wise metrics reveals limitations in the classification of the "Poor" skill level relative to the other categories, as shown in Table \ref*{tab:gb}. This underperformance may be attributed to label inconsistency inherent in the subjective nature of surgical skill assessment. To overcome these limitations, future work will focus on expanding the dataset to encompass a more diverse participant population and enhancing the reliability of ground truth annotations. At present, skill labels are assigned by only two expert raters, potentially constraining the robustness and generalizability of the consensus labels. To establish more reliable and representative ground truth scores, it is imperative to incorporate evaluations from a larger panel of expert raters, thereby reducing subjectivity and improving annotation consistency.

\begin{table}[h]
	\caption{Skill level classification results. }
	\label{tab:gb}
	\centering
	\resizebox{0.8\columnwidth}{!}{
		\begin{tabular}{|c|c|c|c|}
			\hline
			Level    &Precision     & Recall      & F1-score    \\
			\hline
			Poor      & 0.70	    		& 0.68    	&  0.63 	\\
			\hline
			Moderate     & 0.73       & 0.74     	& 0.74   \\
			\hline
			Good    & 0.77     & 0.89   		& 0.83  	\\
			\hline
		\end{tabular}
	}
\end{table}

\section{Conclusion}
This study presents a comprehensive AI-driven framework for the objective assessment of instrument handling skills in microanastomosis procedures. By addressing the inherent limitations of conventional expert-based assessments, the proposed system integrates state-of-the-art computer vision techniques to facilitate standardized, scalable, and reproducible assessment. The system is lightweight and capable of real-time inference, making it suitable for deployment in resource-constrained environments such as surgical training settings. Future work will focus on expanding the dataset and incorporating broader neurosurgical professionals to enhance the reliability of annotation labels, thereby improving the accuracy and robustness of the skill classification model. Ultimately, this research contributes to the ongoing integration of artificial intelligence in surgical education to foster more effective and equitable training outcomes.

\section*{ACKNOWLEDGMENT}
We thank Pokmeng See, MD and Anil Can, MD for assistance with video scoring and rating calibration, Nirav Patel, MD for securing equipment and resources for the project, Emil Petrusa, PhD for support and guidance during the grant funding process.

\begin{comment}

\end{comment}

\bibliographystyle{IEEEtran}
\bibliography{ref}

\end{document}